\documentclass[11pt]{article}

\usepackage[preprint]{acl}

\usepackage{times}
\usepackage{latexsym}

\usepackage[T1]{fontenc}
\usepackage[utf8]{inputenc}

\usepackage{microtype}

\usepackage{inconsolata}

\usepackage{graphicx}

\usepackage{amsmath}
\usepackage{booktabs}
\usepackage{multirow}
\usepackage{siunitx}
\usepackage[most]{tcolorbox}
\usepackage{xcolor}
\usepackage{makecell}
\usepackage[normalem]{ulem}

\newcommand{\textdef}[1]{\textit{#1}}
\newcommand{\word}[1]{``#1''}
\usepackage[colorinlistoftodos]{todonotes}

\title{
    Evaluating Semantic Relation Knowledge \\ 
    through Bidirectional Relation Prediction from Gloss Pairs
}
\author{
  \textbf{Zhihan Cao\textsuperscript{1}} \quad
  \textbf{Adam Nohejl\textsuperscript{2}} \quad
  \textbf{Hitomi Yanaka\textsuperscript{1,2,3}}
\\
  \textsuperscript{1}The University of Tokyo \quad
  \textsuperscript{2}RIKEN \quad
  \textsuperscript{3}Tohoku University
\\
  \small{
    \texttt{zhihancao@g.ecc.u-tokyo.ac.jp} \quad
    \texttt{adam.nohejl@riken.jp} \quad
    \texttt{hyanaka@is.s.u-tokyo.ac.jp}
  }
}
\begin{document}
\maketitle
\begin{abstract}
Semantic relations organise word senses and are fundamental to lexical semantics.
In order to evaluate semantic relation knowledge of large language models~(LLMs), we propose a challenging task in which an LLM predicts the relation between two glosses~(definitions of senses) presented in two different orders.
The predictions for the two orders are evaluated for consistency with the symmetry or asymmetry of semantic relations.
Our task is designed for both monolingual and cross-lingual evaluation, allowing us to examine whether relation knowledge remains accessible when the same senses are verbalised through glosses in different languages.
We construct a dataset of 3,069 sense pairs, with each sense verbalised in English, Greek, Romanian, Turkish, and Japanese.
Experiments with 17 LLMs in zero-shot and five-shot settings reveal generally poor performance.
In each setting, only eight LLMs achieve a macro-averaged F1 score significantly above a random baseline, and only one (zero-shot) and two (five-shot) significantly outperform a neural baseline.
In particular, LLMs often fail to generate bidirectionally consistent predictions for hypernymy and to make correct predictions for holonymy in either direction.
Our results suggest limitations in current LLMs' semantic relation knowledge, both in identifying relations and in respecting asymmetry.
\end{abstract}

\section{Introduction}
\label{sec:intro}
Semantic relations describe how two senses are related~\citep{Cruse_1986}. 
Investigating the extent to which large language models~(LLMs) acquire knowledge of semantic relations provides empirical evidence on whether structured lexical-semantic knowledge can be acquired from unstructured text~\citep{Diera_2026}, and helps improve performance on related tasks~\citep{Torres_2025}.

\begin{figure}[t]
    \centering
    \includegraphics[width=\linewidth]{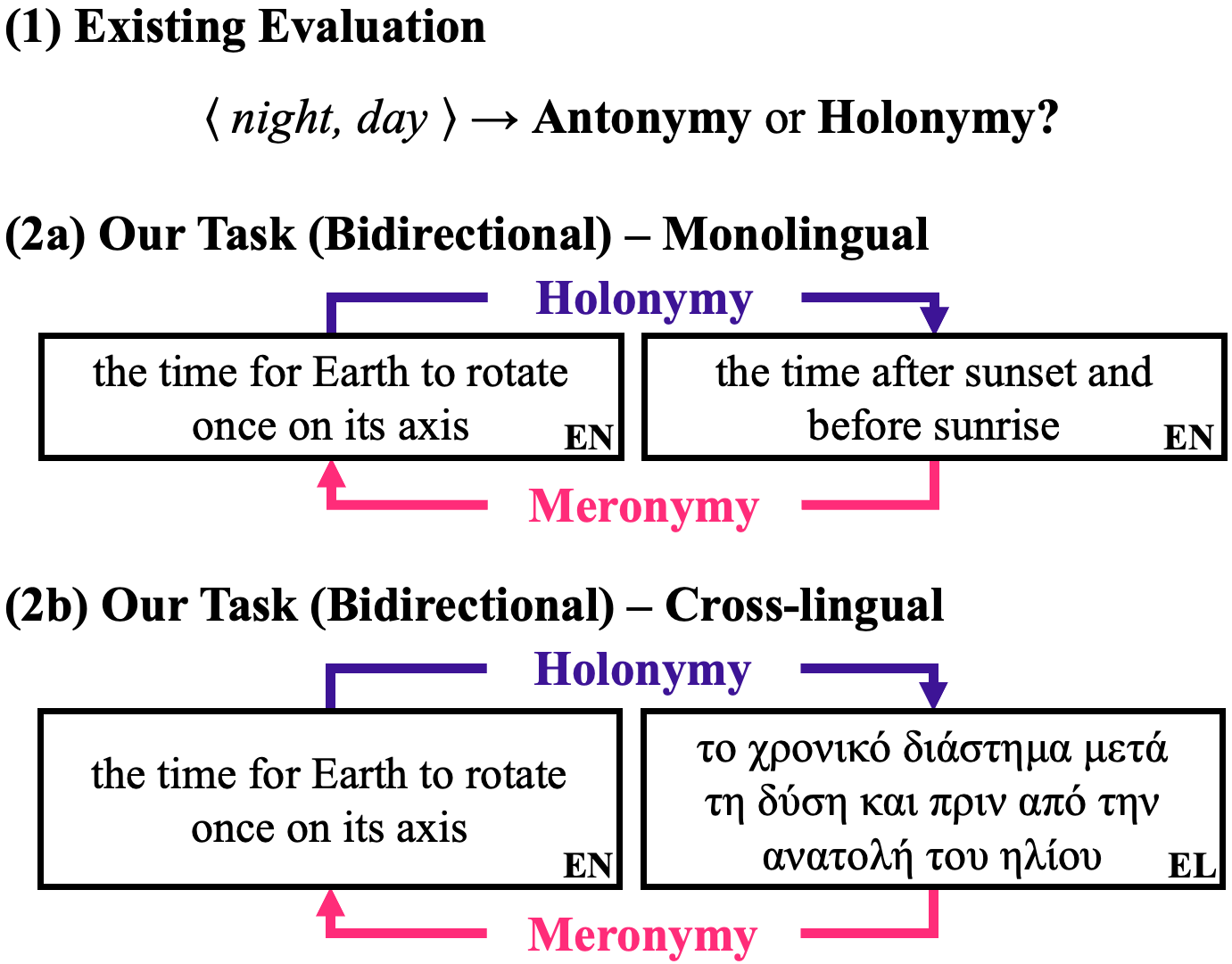}
    \caption{An example of task instances.}
    \label{fig:example}
\end{figure}

Most existing evaluations of LLMs' semantic relation knowledge rely on potentially ambiguous words~\citep{Cao_2024,  Pitarch_2024}.
For example, \word{day} can refer to the time after sunrise and before sunset or the time for Earth to rotate once on its axis; (\word{night}, \word{day}) can therefore instantiate either antonymy or holonymy, as shown in Figure~\ref{fig:example} (1).
Furthermore, the co-occurrence patterns of word pairs differ significantly across semantic relations~\citep{Cao_2025c}, providing LLMs with an additional cue for relation or relata prediction. 
Consequently, model performance is difficult to interpret, as it is unclear which senses underlie the model's predictions and to what extent its performance depends on such statistical cues.

More importantly, existing tasks do not incorporate the properties of semantic relations by construction: they either leave these properties out or measure them separately from relation prediction.
Evaluating these properties is necessary because they are structural constraints that define not only how individual relations behave, but also how different relations are distinguished from or related to one another.
For example, symmetry is an inherent property of antonymy and also distinguishes it from asymmetric relations such as hypernymy.

We argue that an evaluation task for semantic relation knowledge should operate at the sense level and explicitly involve relation properties.
Following the recent attempt by \citet{Cassotti_2026} to use glosses, definitions of senses, as sense representations, we introduce \textbf{bidirectional semantic relation prediction from gloss pairs}.
In our task, given a gloss pair, a model predicts the semantic relation twice, once for each order, as exemplified in Figure~\ref{fig:example} (2a).
Predictions for the two orders are evaluated for consistency with the relation's directional properties.
For example, antonymy pairs should receive antonymy in both directions, whereas holonymy pairs should receive holonymy in one direction and meronymy in the other.

Our task has three advantages.
First, glosses, unlike words that might be polysemous, are associated with a unique sense.
This task therefore requires models to map sense-specific descriptions to semantic relation labels, reducing the mismatch between the target knowledge and its operationalisation.
Second, by filtering out those glosses that contain lexicalisations of the underlying sense, we can limit the extent to which models can exploit co-occurrence cues.
Third, the bidirectional setting assesses whether models capture the directional properties of semantic relations.

Our task supports both monolingual and cross-lingual evaluation, where the two glosses are in the same language and in different languages, respectively, as exemplified in Figure~\ref{fig:example} (2b).
Cross-lingual evaluation matters both semantically and practically.
Semantically, if two related senses can be expressed in multiple languages, their semantic relation should in principle remain constant across linguistic realisations.
Cross-lingual evaluation therefore helps examine whether relation knowledge remains accessible and robust when the same senses are verbalised in different languages.
In practice, LLMs serve users of many languages, yet their performance varies widely across languages~\citep{Nezhad_2024}.
In our dataset, the instances for a given sense pair share the underlying senses and differ mainly in the language of their glosses.
Our evaluation thus largely controls for semantic content across languages, so performance differences are more directly attributable to language.
This provides an empirical basis for developing models with more balanced multilingual ability.

\paragraph{Contributions}
We propose the first evaluation task for semantic relation knowledge that operates at the sense level and incorporates the directional property of each relation, namely symmetry or asymmetry.
We construct a multilingual and cross-lingual dataset that covers antonymy, hypernymy, and holonymy, and contains 3,069 sense pairs, each of which is verbalised in English, Greek, Romanian, Turkish, and Japanese.
This dataset enables controlled comparison across monolingual and cross-lingual settings while keeping the underlying senses fixed.
Together, our task and dataset enable an integrated evaluation of semantic relation knowledge and its cross-lingual robustness.
Our data will be publicly available.

\section{Background and Related Work}
\label{sec:literature}
\subsection{Evaluation Based on Relata Prediction}
Given a target linguistic unit, whether a sense or a word, we call its semantically related linguistic unit a \textit{relatum}.
In evaluation based on relata prediction, an LLM takes an input consisting of a target word and a prompt prescribing a semantic relation and then needs to generate candidate relata of the target word for the prescribed relation.
Consider the input \word{a robin is a type of}.
Here, \word{robin} is the target word and \word{is a type of} prescribes hypernymy.
LLMs are tasked with predicting hypernyms of \word{robin}, such as \word{bird} or \word{animal}.

Initial studies mainly considered hypernymy, examining to what extent models can predict hypernyms of a given target word~\citep {Ettinger_2020, Hanna_2021, Ravichander_2020}.
Recently, \citet{Cao_2024} additionally included more relations and incorporated properties of relations into the evaluation.
However, intra-sentential co-occurrence statistics, such as frequencies and linear orders of pairs, differ significantly both between semantically related and unrelated word pairs and across word pairs that instantiate different semantic relations~\citep{Cao_2025c}.
LLMs might use such co-occurrence statistics, raising the chance of predicting the gold relatum.
It is difficult to disentangle to what extent models rely on co-occurrence or semantic relation knowledge.

\subsection{Evaluation Based on Relation Prediction}
Evaluation based on relation prediction examines, given a pair of semantically related words, whether a model can predict the relation in which they stand.
Earlier research was mainly concerned with monolingual settings~\citep{bless, Weeds_2014, bibless, Shwartz_2016}, while recent literature has extended the task to cross-lingual settings, where the two input elements are presented in different languages~\citep{Glavas_2020,  Pitarch_2024, Gromann_2024}.

Similar to relata prediction, models can also exploit statistical cues in relation prediction~\citep{Kang_2023}.
There is a further limitation that the inputs are pairs of potentially ambiguous words which do not necessarily uniquely determine the semantic relation.
In the existing datasets, nevertheless, each word pair is associated with a single relation.
This creates a mismatch between the information available in the input and the prediction required as the output, reducing the evaluation validity.
It might also unfairly penalise models when the predicted relation is valid but differs from the gold answer, as exemplified by (\word{night}, \word{day}).

Some researchers have gone beyond words by using glosses as input features.
\citet{Cassotti_2024} and \citet{Amorim_2025} used neural classifiers to predict semantic relations between glosses. 
\citet{Cassotti_2026} tasked LLMs with predicting the relation between glosses that belong to the same lexeme.
Many such relations arise through diachronic semantic development, including metaphorical and metonymic extension, and the senses involved might share a degree of semantic continuity~\citep{Traugott_2017}.
Their findings therefore do not necessarily generalise to broader semantic relations where the senses involved need not share a history of semantic development.

\subsection{Our Goal}
Following \citet{Cassotti_2026}, we also use glosses because glosses explicitly describe senses, and using them as inputs allows us to address two methodological concerns.
First, we can substantially reduce the ambiguity inherent in word pair inputs and mitigate the mismatch between the input at the lexical level and the output at the semantic level.
Second, by filtering out glosses that contain the underlying sense's lexicalisations, we ensure that the target word and its relatum never both appear in the input and therefore reduce LLMs' ability to exploit direct co-occurrence cues.
However, we do not limit ourselves to intra-lexemic semantic relations and instead study general semantic relations, namely hypernymy, holonymy, and antonymy.

A further gap concerns how knowledge of relation properties is evaluated.
\citet{Cassotti_2026} primarily assessed whether models predict the correct relation, without incorporating properties of the relations into the task itself.
On the other hand, \citet{Cao_2024} incorporated properties, but they measured properties using separate metrics.
This requires interpretation of multiple scores and does not provide an integrated measure of relation prediction ability and knowledge of properties.
We, in contrast, aim to design the task so that properties of interest, namely symmetry and asymmetry, are incorporated by construction.

\section{Task and Metrics}
\label{sec:task}
%
\iffalse
\begin{table*}[ht]
\centering\small
\begin{tabular}{c|cccc}
\toprule
Outcome
&
$y_{12}=r  \lor y_{21}=\operatorname{inv}(r)$
&
$y_{12}=\operatorname{inv}(y_{21})$
&
Bidirectional Prediction
&
Description
\\
\midrule
$o_1$ & True  & True  & $y_{12}$             & Correct and consistent \\
$o_2$ & True  & False & \textsc{unmatched}   & Partially correct but inconsistent \\
$o_3$ & False & True  & $y_{12}$             & Incorrect but consistent\\
$o_4$ & False & False & \textsc{unmatched}   & Incorrect and inconsistent \\
\bottomrule
\end{tabular}
\caption{
Prediction outcomes for a gloss pair
$(g_1,g_2)$ with gold relation $r$.
$\operatorname{inv}(\cdot)$ is a function that maps a relation to its inverse relation, 
}
\label{tab:prediction_outcomes}
\end{table*}
\fi
%
We propose bidirectional semantic relation prediction from gloss pairs.
In this task, given two glosses, $g_1 \text{ and } g_2$, a model predicts the relation for both $(g_1, g_2)$ and $(g_2, g_1)$, yielding two predictions that we call unidirectional predictions.
The two unidirectional predictions are then combined into a single bidirectional prediction.

Bidirectional predictions allow us to evaluate models' knowledge of the directional properties of semantic relations, namely symmetry and asymmetry.
Antonymy and synonymy are symmetric; reversing the order of an antonymy or synonymy pair does not change the relation instantiated by the pair.
(\word{day}, \word{night}) and (\word{night}, \word{day}) both instantiate antonymy.
Hypernymy and holonymy are asymmetric; reversing the order of their pairs yields the \textdef{inverse relation} of the original relation.
For example, (\word{bird}, \word{animal}) instantiates hypernymy, whereas the reversed pair, (\word{animal}, \word{bird}), instantiates its inverse relation, hyponymy.
For symmetric relations, we consider the inverse relation to be identical to the original relation.

We let $y_{12}$ denote the unidirectional prediction for $(g_1, g_2)$, and $y_{21}$ for $(g_2, g_1)$.
An LLM with knowledge of directional properties is expected to produce mutually consistent predictions in both directions.
Specifically, if $y_{12}$ and $y_{21}$ are inverse relations of each other, then the bidirectional prediction for $(g_1, g_2)$ is $y_{12}$.
In all other cases, the bidirectional prediction is set to a special class \textsc{unmatched}.
A prediction other than \textsc{unmatched} is not necessarily correct but only indicates that the two unidirectional predictions are mutually consistent.
For example, given a hypernymy pair $(g_1, g_2)$, a model might predict hyponymy for $(g_1, g_2)$ and hypernymy for $(g_2, g_1)$, resulting in a consistent but incorrect bidirectional prediction: hyponymy.

We study hypernymy, holonymy, and antonymy\footnote{
    We exclude synonymy because, in monolingual settings, synonymy pairs share a gloss, making the task trivial. 
    In cross-lingual settings, the task largely reduces to translation.
}.
Therefore, for unidirectional prediction, the label space is hypernymy~(HYP), hyponymy~(HPO), holonymy~(HOL), meronymy~(MER), and antonymy~(ANT), and for bidirectional prediction, it is the five relations plus \textsc{unmatched}.

\paragraph{F1 Score}
Our task is a classification task, so we use the F1 score as the primary metric.
For each model, we compute F1 scores at three levels.
The first is for a relation $r$ under a language pair $l$ of the glosses.
We refer to this as the relation-language-specific F1 score and let $\mathcal{F}^l_r$ denote it.
The second is a relation-specific F1 score $\mathcal{F}_r$, obtained by averaging $\mathcal{F}^l_r$ over $l \in L$, where $L$ is the set of language pairs.
The final one measures overall performance $\mathcal{F}$, obtained by further averaging $\mathcal{F}_r$ across $r \in R$, where $R$ is the set of relations.
We report $\mathcal{F}_r$ and~$\mathcal{F}$.

\paragraph{Prediction Outcome Rates}
F1 scores alone do not reveal where prediction errors arise.
To identify the sources of errors, we assign each instance to one of four outcomes based on two criteria: 
whether at least one unidirectional prediction is correct, and whether the two predictions are consistent.
Outcome $o_1$~(correct and consistent) corresponds to a correct bidirectional prediction.
The remaining three outcomes are errors of different types.
In $o_2$~(partially correct but inconsistent), only one of $y_{12}$ and $y_{21}$ is correct, so the bidirectional prediction is \textsc{unmatched}.
In $o_3$~(incorrect but consistent), the two predictions are consistent but both incorrect, so the bidirectional prediction is a relation other than the gold one.
In $o_4$~(incorrect and inconsistent), the two predictions are neither correct nor consistent, so the bidirectional prediction is again \textsc{unmatched}.
We define the \textdef{$o_k$ rate} as the proportion of instances assigned to outcome $o_k$; the four rates sum to one.
As we are primarily interested in the overall pattern, we report the four rates averaged across language pairs and models.

\paragraph{Cross-lingual Coefficient of Variation}
As one of our objectives is to evaluate robustness across gloss language pairs, we need a quantitative measure of such robustness.
We use the coefficient of variation~\citep{Pelabon_2020,Yu_2022,Hu_2025}, defined as the ratio of the standard deviation to the mean.
It quantifies the relative dispersion of performance across the values of a given variable, which in our case are the language pairs.
We define a summary statistic called the cross-lingual coefficient of variation~(CCV).
Formally, CCV is given by
\begin{equation}
    \mathrm{CCV} = \frac{
        \sigma(
            \{\mathcal{F}^{l}\}_{l \in L}
        )
    }{
        \mu(
            \{\mathcal{F}^{l}\}_{l \in L}
        )
    }
    \times100\%
\end{equation}
where $\mathcal{F}^{l}$ is obtained by averaging $\mathcal{F}^{l}_{r}$ across relations, and $\sigma(\cdot)$ and $\mu(\cdot)$ denote the sample standard deviation and mean, respectively.
Although the cross-lingual robustness may differ across semantic relations, we do not compute the coefficient separately for each relation.
This is because 1)~we are interested in overall cross-lingual robustness and 2)~the coefficient might become unstable for relations on which a model’s $\mathcal{F}_r$ is close to zero.
A higher CCV indicates greater cross-lingual variation in performance and therefore lower robustness to gloss language variation.
For interpretation, we use 20\% as a reference threshold in this study\footnote{
    Coefficients of variation are interpreted differently across application domains.
    There is no universally applicable cut-off.
    The cut-off values typically range from 10 to 30\%~\citep[e.g.,][]{Eisen_1988,Atkinson_1998,Pelabon_2020}.
    We use the midpoint of this range as a reference threshold. 
}. 
Models with a CCV above 20\% are considered to exhibit low robustness across language pairs.

\section{Data}
\label{sec:data}
Our task requires a set of semantically related sense pairs, with each sense represented by glosses in multiple languages.
We construct such a dataset using WordNets.
In a WordNet, words sharing a sense are grouped into a set of synonyms~(synset).
Each synset is associated with a gloss and is linked to other synsets through semantic relations.
Specifically, from Cygnet~\citep{Maudslay_Bond_2026}, which brings together more than 40 language-specific WordNets, we
1)~retrieve a set of synsets for which multilingual glosses are available;
2)~retain all hypernymy, holonymy, and antonymy synset pairs;
and 3)~map each synset in a pair to its glosses in target languages, yielding gloss pairs in different language pairs. 

Ideally, our dataset should cover as many languages as possible. 
In practice, however, there is a trade-off between language coverage and the number of available synsets: requiring glosses in more languages reduces the number of synsets shared across all selected languages, which might in turn reduce the statistical precision of model evaluation.

To balance language coverage and synset coverage, we exhaustively enumerate all subsets of languages and count the synsets for which glosses are available in every language in each subset. 
Among subsets containing at least five languages, only four retain more than 7,000 synsets, all of which contain exactly five languages. 
All four subsets include English, Japanese, and Romanian.
The remaining two languages are either German and Uzbek, Turkish and Uzbek, Greek and Uzbek, or Greek and Turkish.

We prefer Greek to German because it increases both the orthographic and genealogical diversity of the selected languages. 
German, like English, uses the Latin script and belongs to the Germanic branch of Indo-European. 
Greek instead uses the Greek script and constitutes the Hellenic branch of the Indo-European language family, complementing the Germanic branch represented by English and the Romance branch represented by Romanian.
Between the two Turkic languages, Uzbek and Turkish, we select Turkish because it improves resource comparability across languages.
The Turkish WordNet was developed together with the Greek and Romanian WordNets as part of BalkaNet~\citep{Stamou_2002}, whereas the Uzbek WordNet was developed separately~\citep{Agostini_2021}.

We further filter out two types of synset pairs that are unsuitable for our purposes.
The first consists of synset pairs with identical glosses, for which the relation cannot be determined from the glosses alone.
The second consists of pairs in which one or both synsets have a gloss containing any of the synset's lexicalisations in at least one language.
Such cases allow models to exploit co-occurrence cues easily.
For example, we exclude all pairs containing the synset \word{chrysanthemum} because its English lexicalisation appears in its English gloss: \word{...herbs having showy brightly colored flower heads of the genera \textit{Chrysanthemum}...}.

In total, we have 3,069 synset pairs:
2,199 nominal and 133 verbal hypernymy pairs; 
659 nominal holonymy pairs; 
and 35 nominal, 21 verbal, and 22 adjectival antonymy pairs.
Each pair is associated with 25 gloss pairs, one for each ordered pair of English, Greek, Romanian, Turkish, and Japanese\footnote{
 The resulting imbalance across relations reflects the distribution of relation instances in WordNets.
 See discussion by~\citet{maziarz_2013}.
 We also present descriptive statistics of glosses by language in Appendix~\ref{app:gloss-stats}.
}.

\section{Experiments}
\label{sec:exp}
\begin{table*}[ht]
\centering\small
\begin{tabular}{l|cccc|cccc}
\toprule
\multicolumn{1}{l}{\multirow{2}{*}{Model}}
& \multicolumn{4}{c}{Zero-shot}
& \multicolumn{4}{c}{Five-shot} \\
\cmidrule(lr){2-5}
\cmidrule(lr){6-9}
\multicolumn{1}{l}{}
& \multicolumn{1}{c}{$\mathcal{F}_{\mathrm{ANT}}$}
& \multicolumn{1}{c}{$\mathcal{F}_{\mathrm{HYP}}$}
& \multicolumn{1}{c}{$\mathcal{F}_{\mathrm{HOL}}$}
& \multicolumn{1}{c}{$\mathcal{F}$}
& \multicolumn{1}{c}{$\mathcal{F}_{\mathrm{ANT}}$}
& \multicolumn{1}{c}{$\mathcal{F}_{\mathrm{HYP}}$}
& \multicolumn{1}{c}{$\mathcal{F}_{\mathrm{HOL}}$}
& \multicolumn{1}{c}{$\mathcal{F}$} \\
\midrule

Tiny Aya (3B)
& \underline{.03} & .12 & .00 & .05
& \textbf{.40}$^\dagger$ & .12\phantom{$^\dagger$} & .02$^\dagger$ & .18$^\dagger$ \\
Aya Expanse (8B)
& \textbf{.45} & .15 & \underline{.16} & \underline{.25}
& \textbf{.47}\phantom{$^\dagger$} & .06\phantom{$^\dagger$} & \underline{.15}\phantom{$^\dagger$} & .23\phantom{$^\dagger$} \\
Aya Expanse~(32B)
& \underline{.24} & .03 & \underline{.11} & .13
& \textbf{.32}$^\dagger$ & .02\phantom{$^\dagger$} & \underline{.08}\phantom{$^\dagger$} & .14\phantom{$^\dagger$} \\

\midrule
EuroMoE (3B)
& \underline{.05} & .00 & .00 & .02
& \underline{.05}\phantom{$^\dagger$} & .00\phantom{$^\dagger$} & .00\phantom{$^\dagger$} & .02\phantom{$^\dagger$} \\
EuroLLM (9B)
& .00 & .04 & \underline{.13} & .06
& \underline{.22}$^\dagger$ & .04\phantom{$^\dagger$} & \underline{.11}\phantom{$^\dagger$} & .12$^\dagger$ \\
EuroLLM (22B)
& \underline{.14} & .08 & .00 & .07
& \underline{.22}$^\dagger$ & .05\phantom{$^\dagger$} & \underline{.09}$^\dagger$ & .12$^\dagger$\\

\midrule
Marco-Nano (8B)
& .00 & .00 & .00 & .00
& \underline{.15}$^\dagger$ & .00\phantom{$^\dagger$} & .00\phantom{$^\dagger$} & .05$^\dagger$ \\
Marco-Mini (17B)
& \underline{.09} & .01 & .00 & .03
& \textbf{.29}$^\dagger$ & .24$^\dagger$ & .01$^\dagger$ & .18$^\dagger$ \\

\midrule
Gemma 4 (12B)
& \underline{.05} & .05 & .01 & .04
& \underline{.31}$^\dagger$ & .00\phantom{$^\dagger$} & .01\phantom{$^\dagger$} & .11\phantom{$^\dagger$} \\
Gemma 4 (26B)
& \underline{.28} & \underline{.42} & .00 & \underline{.23}
& \textbf{.31}\phantom{$^\dagger$} & \underline{.36}\phantom{$^\dagger$} & .00\phantom{$^\dagger$} & \underline{.22}\phantom{$^\dagger$} \\
Gemma 4 (31B)
& \textbf{.57} & \textbf{.71} & .04 & \underline{.44}
& \textbf{.67}$^\dagger$ & \textbf{.69}\phantom{$^\dagger$} & \underline{.06}\phantom{$^\dagger$} & .\textbf{47}$^\dagger$ \\

\midrule
Qwen3.5 (9B)
& \underline{.28} & .01 & .00 & .10
& \textbf{.28}\phantom{$^\dagger$} & .22$^\dagger$ & \underline{.05}$^\dagger$ & \underline{.18}$^\dagger$ \\
Qwen3.5 (27B)
& \textbf{.36} & \textbf{.60} & .02 & \underline{.33}
& \textbf{.54}$^\dagger$ & \textbf{.66}\phantom{$^\dagger$} & \underline{.17}$^\dagger$ & \textbf{.46}$^\dagger$ \\
Qwen3.5 (35B)
& \textbf{.57} & \textbf{.60} & .01 & \underline{.39}
& \textbf{.41}\phantom{$^\dagger$} & \underline{.39}\phantom{$^\dagger$} & \underline{.05}$^\dagger$ & \underline{.28}\phantom{$^\dagger$} \\
Qwen3.5 (122B)
& \textbf{.55} & \underline{.50} & .01 & \underline{.35}
& \textbf{.42}\phantom{$^\dagger$} & \underline{.38}\phantom{$^\dagger$} & \underline{.06}$^\dagger$ & \underline{.29}\phantom{$^\dagger$} \\

\midrule
GPT-5.6 Luna
& \boxed{\textbf{.72}} & \textbf{.72} & .01 & \underline{.48}
& \textbf{.46}\phantom{$^\dagger$} & \underline{.51}\phantom{$^\dagger$} & .01\phantom{$^\dagger$} & \underline{.33}\phantom{$^\dagger$} \\
GPT-5.6 Terra
& \textbf{.44} & \textbf{.68} & \underline{.18} &\textbf{.43}
& \textbf{.48}\phantom{$^\dagger$} & \underline{.54}\phantom{$^\dagger$} & \underline{.20}\phantom{$^\dagger$} & \underline{.41}\phantom{$^\dagger$} \\

\midrule
Random Baseline
& .00 & .26 & .01 & .09
& .00\phantom{$^\dagger$} & .26\phantom{$^\dagger$} & .01\phantom{$^\dagger$}& .09\phantom{$^\dagger$} \\
Neural Baseline
& .24 & .49 & \boxed{.31} & .35
& .24\phantom{$^\dagger$} & .49\phantom{$^\dagger$} & \boxed{.31}\phantom{$^\dagger$} & .35\phantom{$^\dagger$} \\

\midrule
\midrule
$\#$ LLMs Outperforming Random Baseline
& 15
& 7
& 4
& 8
& 17\phantom{$^\dagger$}
& 7\phantom{$^\dagger$}
& 10\phantom{$^\dagger$}
& 8\phantom{$^\dagger$} \\
$\#$ LLMs Outperforming Neural Baseline
& 7
& 5
& 0
& 1
& 12\phantom{$^\dagger$}
& 2\phantom{$^\dagger$}
& 0\phantom{$^\dagger$}
& 2\phantom{$^\dagger$} \\
$\#$ LLMs Benefiting from Exemplars
& -- 
& -- 
& -- 
& --
& 9\phantom{$^\dagger$}
& 2\phantom{$^\dagger$}
& 7\phantom{$^\dagger$}
& 8\phantom{$^\dagger$} \\
\bottomrule
\end{tabular}

\caption{
    F1 scores under zero-shot and five-shot settings for different relations.
    Underlining indicates performance significantly higher than the random baseline but not the neural baseline.
    Boldface indicates performance significantly higher than the neural baseline.
    Boxing indicates performance that is significantly higher than all other models.
    $^\dagger$ indicates a significant improvement from zero-shot to five-shot.
}
\label{tab:f1}
\end{table*}
We evaluate 15 open-source instruction-tuned models that have broad multilingual coverage supporting English, Greek, Japanese, Romanian, and Turkish\footnote{
    We present the details of the models in Appendix~\ref{app:model_details}.
}:
Tiny Aya~\citep[3B, dense;][]{tinyaya},
Aya Expanse~\citep[8B and 32B, dense;][]{ayaexpanse}, 
EuroMoE~\citep[3B, mixture-of-experts;][]{euromoe},
EuroLLM~\citep[9B and 22B, dense;][]{eurollm},
Marco-Nano and Marco-Mini~\citep[8B and 17B, respectively, mixture-of-experts;][]{marco-moe}, 
Gemma 4~\citep[12B and 31B, dense; 26B, mixture-of-experts;][]{gemma4}, 
and Qwen3.5~\citep[9B and 27B, dense; 35B and 122B, mixture-of-experts;][]{qwen3.5}. 
For proprietary models, we include GPT-5.6 Luna and GPT-5.6 Terra~\citep{gpt56}.
As not all models support reasoning, we evaluate all models without reasoning to ensure comparability.
The parameter counts of the models range from a few billion to more than one hundred billion, including dense and mixture-of-experts architectures. 
This diversity in model scale and architecture allows us to examine whether the observed tendencies generalise beyond any particular model.

We evaluate LLMs under zero-shot and five-shot settings. 
Each model is instructed to output a single digit~(1–5) denoting one of the five unidirectional labels.
In order to ensure comparability, we use greedy decoding and LLMs only need to predict one token.
Any other output is scored as incorrect\footnote{
    We present the system prompt in Appendix~\ref{app:sysprom}.
    Replacing the relation terms in the prompt with plain descriptions (e.g., replacing \word{meronym} with \word{is a part of}) does not significantly change performance in the preliminary experiments.
}.

In the five-shot setting, we sample one exemplar gloss pair for each of hypernymy, hyponymy, holonymy, meronymy, and antonymy.
These five exemplars are used across instances and excluded from the evaluation data in both settings.
We match the language pair of the exemplars to that of the input instance.
For example, if the first gloss is English and the second gloss is Turkish, then all five exemplars use English for the first gloss and Turkish for the second.

We consider two baselines, both of which generate unidirectional predictions independently like LLMs.
\textdef{The random baseline} samples a relation according to the empirical relation distribution in the dataset and provides us with a chance-level reference.
\textdef{The neural baseline} is a neural classifier that takes gloss embeddings produced by multilingual-E5-large~\citep{e5} as input.
We ensure that the neural baseline receives no explicit supervision about the directional properties of these relations, so its performance indicates the extent to which both relation information and directional properties are recoverable from the glosses\footnote{
    We provide details of the two baselines in Appendix~\ref{app:baselines}.
}.

We use the Wilcoxon signed-rank test, with a significance level of 0.01, to compare model performance. 
For each relation, a model is characterised by 25 F1 scores, one for each language pair. 
Across the three relations, this yields 75 F1 scores per model.
We perform three types of paired comparisons: 
1) pairwise comparisons across models for each semantic relation;
2) pairwise comparisons across models for overall performance; and 3) comparisons between zero-shot and five-shot performance.
This results in 1,436 statistical tests in total, which increases the risk of Type I errors due to multiple testing.
We reduce the risk by applying the Benjamini--Yekutieli correction~\citep{benjamini_2001}.
It controls the expected proportion of incorrectly rejected null hypotheses among all rejected hypotheses at our significance level.

\section{Results}
\label{sec:results}
\subsection{F1 Scores}
In Table~\ref{tab:f1}, we report $\mathcal{F}_\mathrm{ANT}$, $\mathcal{F}_\mathrm{HYP}$, $\mathcal{F}_\mathrm{HOL}$, and $\mathcal{F}$ for all LLMs and the two baselines\footnote{
    For the two baselines, we report the same evaluation results for both settings for ease of comparison.
}
under zero-shot and five-shot settings.
Unless otherwise specified, we discuss only statistically significant differences in F1 scores.
Overall, LLMs perform poorly on our task.
In each setting, only eight of the seventeen LLMs outperform the random baseline.
Even fewer outperform the neural baseline: one in the zero-shot setting and two in the five-shot setting.
No LLM achieves an $\mathcal{F}$ above 0.5.

The baselines' advantage holds for holonymy and hypernymy but not for antonymy.
For holonymy, no LLMs outperform the neural baseline in either setting.
For hypernymy, fewer than half of the LLMs outperform the two baselines in either setting.
For antonymy, seven and twelve LLMs outperform the neural baseline in the zero-shot and five-shot settings, respectively; at least fifteen LLMs outperform the random baseline in both settings.

Overall, neither additional exemplars nor greater model scale consistently improves performance on this task.
Only eight LLMs achieve a higher $\mathcal{F}$ in the five-shot setting than in the zero-shot setting.
Within the same family, the larger variants do not necessarily outperform their smaller counterparts.
For example, Aya Expanse~(32B) has lower $\mathcal{F}$ than Aya Expanse~(8B) in both settings.

\subsection{Prediction Outcome Rates}
\begin{table}[ht]
\centering\small
\begin{tabular}{cc|rrrr}
\toprule
Relation & S & $o_1$ & $o_2$ & $o_3$ & $o_4$ \\
\midrule
\multirow{2}{*}{ANT} 
& 0 & 50\% & 21\% & 1\%  & 28\% \\
& 5 & 73\% & 15\% & 1\%  & 11\% \\
\midrule

\multirow{2}{*}{HYP} 
& 0 & 22\% & 39\% & 13\%  & 27\% \\
& 5 & 18\% & 44\% & 17\%  & 21\% \\
\midrule

\multirow{2}{*}{HOL} 
& 0 & 3\% & 18\% & 27\%  & 52\% \\
& 5 & 4\% & 26\% & 24\% & 46\% \\
\bottomrule
\end{tabular}

\caption{
    Model-average prediction outcome rates. 
    S refers to the setting, with 0 indicating zero-shot and 5 indicating five-shot.
}
\label{tab:rates}
\end{table}
Table~\ref{tab:rates} shows the outcome rates averaged across LLMs, broken down by relation and setting.
The three relations exhibit distinct outcome patterns, which remain largely
stable across the zero-shot and five-shot settings.
Antonymy is the only relation for which correct bidirectional predictions
dominate, with $o_1$~(correct and consistent) rates of at least 50\% in both settings.
For hypernymy, the most frequent outcome (at least 39\% across settings) is $o_2$~(partially correct).
LLMs often identify the relation in one direction but not in the other.
For holonymy, neither unidirectional prediction is often correct, as shown by the high $o_4$~(incorrect and inconsistent) rates of at least 46\%.

\begin{table}[th]
    \centering
    \small
        \begin{tabular}{cc|rrrrr}
        \toprule
        Gold & 
        S & ANT & HOL & MER & HYP & HPO \\
        \midrule
        \multirow{2}{*}{HOL}
        & 0 & 19\% & -- & 6\% & 59\% & 15\% \\
        & 5 & 26\% & -- & 14\% & 44\% & 17\% \\
        
        \midrule
        \multirow{2}{*}{HYP}
        & 0 & 52\% & 9\% & 4\% & -- & 34\% \\
        & 5 & 64\% & 8\% & 6\% & -- & 23\% \\
        
        \bottomrule
    \end{tabular}

    \caption{Predicted relation distributions among consistent but incorrect predictions, averaged across LLMs for holonymy and hypernymy by setting. 
    Columns indicate predicted relations.}
    \label{tab:confusion}
\end{table}
Holonymy and hypernymy also show considerably higher $o_3$~(incorrect but consistent) rates than antonymy~(13--27\% vs. 1\%).
This indicates that, for these two relations, the two unidirectional predictions are often mutually consistent but are different from the gold one, suggesting confusion with other relations.
In Table~\ref{tab:confusion}, we break down these consistent errors by the predicted relation.
Holonymy is most frequently confused with hypernymy (at least 44\% in both settings), followed by antonymy (at least 19\%), whereas hypernymy is most frequently confused with antonymy (at least 52\%), followed by hyponymy (at least 23\%).

\subsection{Cross-lingual Coefficients of Variation}
\begin{table}[ht]
    \centering\small
        \begin{tabular}{l|cc}
    \toprule
    Model
    & \multicolumn{1}{c}{Zero-shot}
    & \multicolumn{1}{c}{Five-shot} \\
    \midrule
        %Tiny Aya (3B)     & \textbf{48.5} & 15.6 \\
        %Aya Expanse (8B)  & 14.4 & 10.2 \\
        %Aya Expanse (32B)  & \textbf{31.2} & 12.4 \\
        %\midrule 
        %EuroMoE (3B) & \phantom{0}8.7 & \phantom{0}3.5\\
        %EuroLLM (9B)   & \textbf{31.1} & \textbf{25.2}\\
        %EuroLLM (22B)  & \textbf{22.9} & \textbf{29.6}\\
        %\midrule
        %Marco-Nano (8B) & -- & \textbf{33.6} \\
        %Marco-Mini (17B) & \textbf{89.6} & 11.9\\
        %\midrule
        %Gemma 4 (12B)  & \textbf{45.1} & 51.1\\
        %Qwen3.5 (9B)  & 17.9 & 13.8 \\
        Gemma 4 (26B)  & 18.9 & 14.5\\
        Gemma 4 (31B)  & \phantom{0}9.1 & \phantom{0}9.8\\
        Qwen3.5 (27B)  & 12.8 & 10.1\\
        Qwen3.5 (35B)  & 10.3 & 12.8\\
        Qwen3.5 (122B)  & 11.5 & 14.0\\
        GPT-5.6 Luna  & \phantom{0}5.9 & 15.8\\
        GPT-5.6 Terra  & 12.6 & \phantom{0}9.6\\
    \bottomrule
    \end{tabular}
    \caption{
        CCV~(\%) of LLMs by setting. 
    }
    \label{tab:ccv}
\end{table}
In Table~\ref{tab:ccv}, we report the CCV of LLMs outperforming the random baseline in both settings\footnote{
    The full results can be found in Appendix~\ref{app:ccv_all}.
}.
Such LLMs have CCV values below our reference threshold of 20\%, indicating that their performance varies little across language pairs.
From the zero-shot to the five-shot setting, CCV decreases for three and increases for four out of the seven LLMs, suggesting that exemplars do not necessarily improve robustness across language pairs.

\section{Discussion}
\label{sec:dis}
We first discuss factors behind LLMs' generally poor performance and then turn to their cross-lingual performance.
Several factors may contribute, including the information available in the data and LLMs' knowledge of semantic relations.
Among them, insufficient information in the gloss pairs is unlikely to be the main cause.
Our neural baseline performs relatively well, consistent with prior work~\citep{Cassotti_2026, Amorim_2025}, suggesting that the gloss pairs do contain useful information for identifying their relations.

The LLMs studied often fail to produce bidirectionally consistent predictions when the order of the gloss pair is reversed.
LLMs are known to be sensitive to input order; this sensitivity might partially explain inconsistency~\citep{Liusie_2024, Pezeshkpour_2024}.
However, it does not fully explain the fact that the LLMs studied tend to produce more bidirectionally inconsistent predictions for hypernymy than for antonymy.
A closer look at hypernymy pairs for which only one of the two unidirectional predictions is correct shows that, averaged across models, 68\% (zero-shot) and 74\% (five-shot) receive the same label in both orders (i.e., hypernymy for both or hyponymy for both).
That is, the models tend to assign the same label regardless of the gloss order, as if hypernymy were symmetric. 
Such tendencies are hard to attribute to general order sensitivity, which would make predictions vary with order rather than remain unchanged.
Therefore, the bidirectional inconsistency may also stem from LLMs' limited knowledge of the directional properties of semantic relations.
This aligns with the previous finding that relation directionality is only weakly encoded in LLMs' representation space~\citep{Cao_2026}.

Limited knowledge of semantic relations may also explain LLMs' confusion among relations.
Among holonymy instances with consistent but incorrect predictions, more than 60\% are predicted as either hypernymy or hyponymy.
The similarity between holonymy and hypernymy has been established in the theoretical and cognitive literature~\citep{Cruse_1986, Winston_1987, Joosten_2010}, and confusion between these relations has also been observed for other LLMs on relata prediction tasks~\citep{Cao_2024}.
LLMs may fail to capture distinctive characteristics of these two relations and have difficulty distinguishing them.

We also observe that LLMs frequently misclassify hypernymy pairs, and to a lesser degree holonymy pairs, as antonymy.
This confusion, together with LLMs' high performance on antonymy, is consistent with the antonymy bias reported in relata prediction~\citep{Cao_2024}.
In relata prediction, this bias has been attributed to the strong lexical co-occurrence of antonyms~\citep{Cao_2025c}.
However, this explanation is less applicable to our results, where sense pairs are represented by their glosses rather than their word forms, and models cannot directly exploit the co-occurrence of underlying senses' lexicalisations as additional cues. 
We therefore suggest that the bias in relation prediction may stem, at least in part, from the semantic characteristics of antonymy itself.
Antonyms are alike in most respects and contrast along a single salient dimension~\citep{Cruse_1986, Murphy_2003}. 
Similarity and contrast may also hold for hypernymy and holonymy, but are not sufficient to identify these two relations, as they are additionally characterised by generalisation and part-whole inclusion, respectively.
If an LLM recognises that the two underlying senses are similar yet contrasting but fails to identify these additional characteristics, antonymy becomes the most plausible among the five semantic relations.
On this view, the antonymy bias in the two tasks may be similar on the surface but driven by different mechanisms: primarily lexical co-occurrence in relata prediction, and minimal semantic difference in relation prediction.

LLMs that outperform the random baseline show relative cross-lingual robustness in our task.
This contrasts with previous findings that cross-lingual performance can be sensitive to factors such as linguistic similarity and resource availability~\citep{Nezhad_2024}.
One possible explanation lies in the characteristics of glosses.
On average, the glosses used contain 10–13 words in the five languages, and 74–84\% have nominal heads, with most of the remainder being verb-headed phrases whose form follows each language's dictionary conventions\footnote{
    Details can be found in Appendix~\ref{app:gloss-stats}.
}.
We hypothesise that this shared style across the five languages reduces cross-lingual differences, making the task similarly difficult across language pairs.

\section{Conclusion}
\label{sec:con}
We propose bidirectional semantic relation prediction from gloss pairs, a challenging task for evaluating LLMs' multilingual and cross-lingual knowledge of semantic relations.
After careful data cleaning, we obtain 3,069 sense pairs whose glosses are available in English, Greek, Romanian, Turkish, and Japanese.
Our evaluation experiments on 15 open-source LLMs and 2 proprietary LLMs under both zero-shot and five-shot settings show that they generally perform poorly.
Although the LLMs achieve good performance for antonymy, they often fail to generate bidirectionally consistent predictions for hypernymy and to make correct predictions for holonymy in either direction.
LLMs that outperform the random baseline show relative robustness across gloss language pairs.
Our results reveal the weaknesses shared across the LLMs studied, calling into question whether they possess reliable knowledge of semantic relations.

\section{Limitations}
Our study covers three semantic relations, whose distribution is highly imbalanced.
Estimates of $\mathcal{F}_\mathrm{ANT}$ are considerably noisier than those for the other two relations.
Because we exclude synonymy for the reasons given in Section~\ref{sec:task}, antonymy is the only symmetric relation in our data.
Our findings concerning symmetric relations rest on a small subset. Our data are also dominated by nominal pairs, and our results should accordingly be read as pertaining primarily to nouns. 
A more balanced dataset covering a wider range of semantic relations is necessary for a more comprehensive evaluation.

Our five languages cover only a small portion of the world's languages. 
The Japanese glosses are translated from their English originals.
The extent to which our cross-lingual results reflect properties of the languages themselves rather than of the translation process therefore remains open. 

Given our aim of evaluating the semantic relation knowledge of currently available LLMs, we neither enable reasoning nor fine-tune models on our task. 
Enabling reasoning would introduce reasoning ability as an additional factor and make it harder to interpret the performance with respect to relation knowledge.
Fine-tuning is beyond our scope, as we aim to evaluate knowledge acquired by current models.

We provide neither a human upper bound nor a qualitative analysis of model predictions.
The former would help us establish how difficult our task is in absolute terms; the latter would help us identify which semantic or syntactic properties of glosses are useful for models.
Constructing a human upper bound is, however, difficult in practice, as the cross-lingual instances require annotators proficient in both languages of each language pair (e.g., Greek and Japanese), and a human upper bound for the monolingual instances alone would not characterise the entire task.
In the absence of a human upper bound, the neural baseline offers a reference point, indicating that the relations are recoverable from the gloss pairs to a non-trivial extent.

WordNet glosses are publicly available and might have been seen during pretraining.
We cannot exclude prior exposure to the glosses themselves, which could inflate the scores we report.

Our study is diagnostic: it shows that LLMs have limited knowledge of semantic relations, particularly of their directional properties.
However, it does not address this weakness, which matters in practice. 
LLMs have been used to extend existing lexical resources~\citep{Augello_2026}, and predictions not grounded in correct knowledge may introduce errors. 
We leave this for future work.

\bibliography{custom}

\appendix
\label{sec:appendix}
\section{Model Information}
\label{app:model_details}
\begin{table*}[ht]
\centering
\footnotesize
\renewcommand{\arraystretch}{1.08}
\begin{tabular}{
    l|ccc
}
\toprule
Model
& Hugging Face ID
& \# Languages
& \# Parameters (B) \\
\midrule

Tiny Aya (3B)
& \texttt{CohereLabs/tiny-aya-global}
& 70$^\dagger$
& 3.35\\
Aya Expanse (8B)
& \texttt{CohereLabs/aya-expanse-8b}
& 23$^\dagger$
& 8\\
Aya Expanse (32B)
& \texttt{CohereLabs/aya-expanse-32b}
& 23$^\dagger$
& 32\\
\midrule

EuroMoE (3B)
& \texttt{utter-project/EuroMoE-2.6B-A0.6B-Instruct-2512}
& 35$^\dagger$
& 2.6\\
EuroLLM (9B)
& \texttt{utter-project/EuroLLM-9B-Instruct-2512}
& 35$^\dagger$
& 9\\
EuroLLM (22B)
& \texttt{utter-project/EuroLLM-22B-Instruct-2512}
& 35$^\dagger$
& 22\\
\midrule

Marco-Nano (8B)
& \texttt{ATH-MaaS/Marco-Nano-Instruct}
& 29$^\dagger$ 
& 8\\
Marco-Mini (17B)
& \texttt{ATH-MaaS/Marco-Mini-Instruct}
& 29$^\dagger$ 
& 17.3\\
\midrule

Gemma 4 (12B)
& \texttt{google/gemma-4-12B-it}
& 35+
& 11.95\\
Gemma 4 (26B)
& \texttt{google/gemma-4-26B-A4B-it}
& 35+
& 25.2\\
Gemma 4 (31B)
& \texttt{google/gemma-4-31B-it}
& 35+
& 30.7\\
\midrule

Qwen3.5 (9B)
& \texttt{Qwen/Qwen3.5-9B}
& 201 
& 9 \\
Qwen3.5 (27B)
& \texttt{Qwen/Qwen3.5-27B}
& 201 
& 27\\
Qwen3.5 (35B)
& \texttt{Qwen/Qwen3.5-35B-A3B}
& 201 
& 35\\
Qwen3.5 (122B)
& \texttt{Qwen/Qwen3.5-122B-A10B}
& 201 
& 122 \\

\bottomrule
\end{tabular}

\caption{
Details of the open-source models evaluated in this study.
\# Languages indicates the number of languages the model supports; \# Parameters~(B) indicates the total number of parameters in billions.
$^\dagger$ indicates that the developers provide a complete enumeration of the supported languages.
}
\label{tab:model_details}
\end{table*}
The details of each open-source model are given in Table~\ref{tab:model_details}.
We accessed all open-source and proprietary models in July 2026.
The open-source models were loaded in bfloat16 precision.

\section{System Prompt}
\label{app:sysprom}
We use the following system prompt.
\begin{quote}
\small
\textbf{System Prompt}

Determine how Sense B is semantically related to Sense A based solely on their glosses.

\begin{description}
    \item[\texttt{1 =}] Sense B is a hypernym of Sense A 
    \item[\texttt{2 =}] Sense B is a hyponym of Sense A 
    \item[\texttt{3 =}] Sense B is an antonym of Sense A 
    \item[\texttt{4 =}] Sense B is a holonym of Sense A 
    \item[\texttt{5 =}] Sense B is a meronym of Sense A 
\end{description}

Output exactly one digit: \texttt{1}, \texttt{2}, \texttt{3}, \texttt{4},
or \texttt{5}, with no additional text, including reasoning,
explanations, or any other content.
\end{quote}

Terms such as \word{hypernymy} and \word{holonymy} might be rare in the training corpora, making models perform poorly on holonymy pairs.
In preliminary experiments, we compared the system prompt above with the following alternative system prompt, which describes relations in natural language instead of terms.
\begin{quote}
\small
\textbf{Alternative System Prompt}

Determine how Sense B is semantically related to Sense A based solely on their glosses.

\begin{description}
    \item[\texttt{1 =}] Sense B is a broader category that includes Sense A
    \item[\texttt{2 =}] Sense B is a narrower category within Sense A
    \item[\texttt{3 =}] Sense B is opposite in meaning to Sense A
    \item[\texttt{4 =}] Sense B is a whole that contains Sense A as a part
    \item[\texttt{5 =}] Sense B is a part of the whole denoted by Sense A
\end{description}

Output exactly one digit: \texttt{1}, \texttt{2}, \texttt{3}, \texttt{4},
or \texttt{5}, with no additional text, including reasoning,
explanations, or any other content.
\end{quote}

We compared the accuracy scores of Gemma 4~(26B) using the two system prompts in the preliminary experiments and found no significant difference in their performance~(0.434 vs. 0.437, Wilcoxon signed-rank test, $\alpha$ = 0.01).
We therefore use the former.

\section{Glosses}
\begin{table*}[ht]
\centering\small
\begin{tabular}{l|rrrrrr}
\toprule
 & \multirow{2}{*}{Length (words)} 
 & \multirow{2}{*}{Finite clauses (\%)}
 & \multicolumn{4}{c}{Syntactic head (\%)}\\
\cmidrule(lr){4-7}
Language & &  & Nom. & Verb. & Adj./Adv. & Other\\
\midrule
English & 10.7 & 1.9 & 79.2 & 17.3 & 1.6 & 1.8 \\
Greek & 12.9 & 19.6 & 77.4 & 18.4 & 4.0 & 0.2  \\
Romanian & 12.8 & 3.6 & 81.2 & 16.8 & 1.9 & 0.1 \\
Turkish & 10.2 & 3.8 & 73.7 & 19.2 & 6.7 & 0.4 \\
Japanese & 11.1 & -- & 83.6 & 15.7 & 0.7 & 0.0 \\
\bottomrule
\end{tabular}
\caption{Descriptive statistics of the glosses.}
\label{tab:gloss-stats}
\end{table*}

\subsection{Descriptive Statistics}
\label{app:gloss-stats}
To characterise the glosses used in our experiments, we computed descriptive statistics of their length and syntactic form in each language.
All glosses were tokenised, part-of-speech tagged, and dependency parsed with Stanza~\citep[version 1.15.0;][]{qi-etal-2020-stanza}, using the default models for each language.

We measure length as the number of UD word tokens, excluding punctuation.
We prefer this unit to whitespace-separated tokens, which are not available for Japanese and conflate several words in agglutinative Turkish, and to characters, which are not comparable across writing systems.

We describe each gloss by the universal part-of-speech tag of its syntactic root, grouped into nominal (\textsc{noun}, \textsc{propn}, \textsc{pron}, \textsc{num}), verbal (\textsc{verb}, \textsc{aux}), adjectival or adverbial (\textsc{adj}, \textsc{adv}), and other heads.
For glosses consisting of more than one sentence, the root of the first sentence is used.

To distinguish glosses that are sentences from glosses that are phrases, we determined whether each gloss contains a finite independent clause, using UD morphological features.
A gloss of several sentences counts as containing a finite clause if any of its sentences does.
Because Japanese UD does not annotate finiteness, this measure is not reported for Japanese.

Table~\ref{tab:gloss-stats} shows the results.
Glosses are of similar length in all five languages, with means between 10.2 and 12.9 words.
In every language, most glosses have a nominal head (73.7--83.6\%). 
Glosses with a verbal head usually follow the citation conventions of each language: infinitives in English and Romanian, infinitives in \textit{-mak/-mek} in Turkish, dictionary forms in Japanese, and finite forms in Greek.
The last point accounts for the higher proportion of finite clauses among Greek glosses (19.6\%), compared with 1.9--3.8\% in English, Romanian, and Turkish.
Glosses in all five languages are thus predominantly phrasal, and they are similar in length and syntactic form across languages.

\subsection{Gloss Similarity}
\label{app:glsim}
\begin{figure*}[ht]
    \centering
    \includegraphics[width=.95\linewidth]{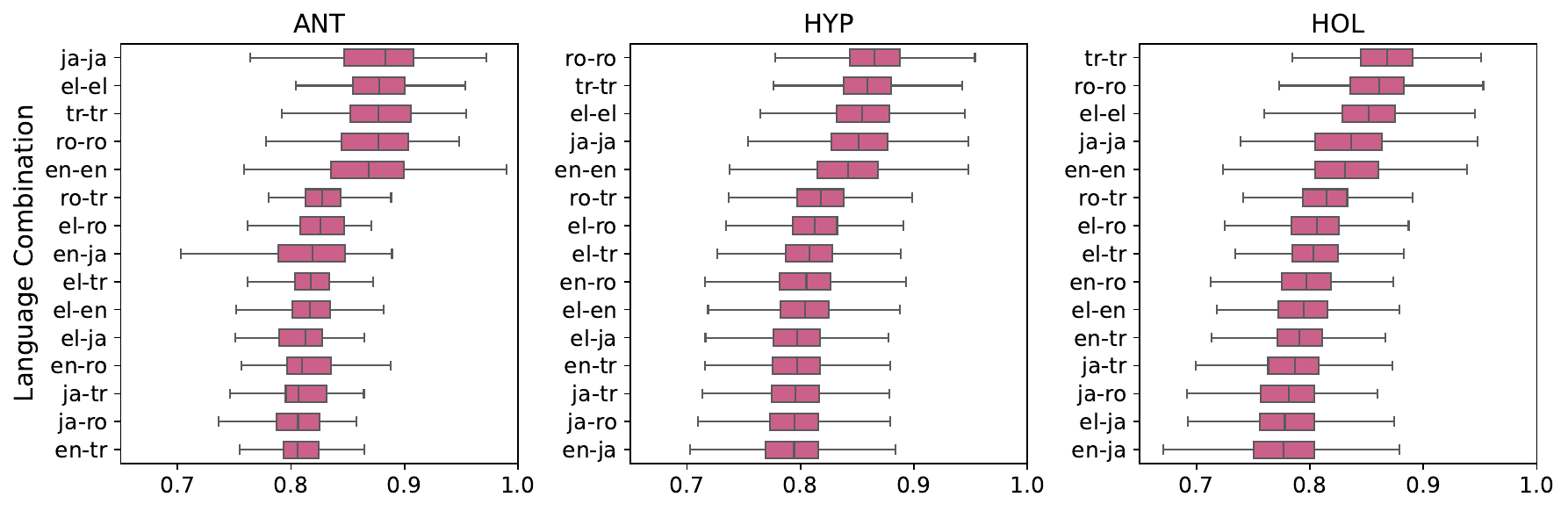}
    \caption{
    Cosine similarity of gloss pairs across language pairs, by relation.
    }
    \label{fig:gloss_sim}
\end{figure*}
In order to characterise our dataset, we evaluate the similarity of gloss pairs.
We embed glosses into vectors using multilingual-E5-large~\citep[mE5-large;][]{e5} and calculate the cosine similarity of gloss embeddings for different relations and language pairs.
We present the distribution of cosine similarity across language pairs for each relation in~Figure~\ref{fig:gloss_sim}, with language pairs ordered by decreasing median similarity.

Across relations, the five monolingual pairs have the highest similarity.
Overall median cosine similarity is highest for antonymy (0.84), followed by hypernymy (0.82) and holonymy (0.81).
However, gloss similarity does not appear to consistently predict model performance.
In both settings, more than 90\% of the model-level point-biserial correlation coefficients between gloss cosine similarity and prediction correctness fall between $-0.03$ and $0.15$.

\section{Baselines}
\label{app:baselines}
We use a bilinear neural classifier as our neural baseline.
Given two gloss embeddings $x$ and $y$, the neural baseline computes~$$h=\operatorname{ReLU}(Ux+Vy+x^TWy+b),$$
where $U$, $V$, $W$ and $b$ are learnable parameters.
The classifier then maps $h$ to a distribution over hypernymy, hyponymy, antonymy, holonymy, and meronymy. 
We use a hidden size of 128, a learning rate of 0.001, and a batch size of 2,048.

During training, we randomly reverse the order of some sense pairs to create hyponymy and meronymy pairs.
When the order is reversed, hypernymy and holonymy are relabelled as hyponymy and meronymy, respectively.
The order of a sense pair is shared across all language pairs.
We ensure that a sense pair does not appear in two orders.
Therefore, the neural baseline receives no explicit supervision about the directional properties of these relations. 
We train the neural baseline using class-weighted cross-entropy loss and the AdamW optimiser.
For each fold, we split the data such that neither sense in a training pair occurs in any test pair.
This prevents the neural baseline from simply learning relations that frequently co-occur with particular glosses~\citep[cf. lexical memorisation;][]{Levy_2015}.
At evaluation time, we apply the neural baseline to both orders of each gloss pair and combine the two predictions using the same procedure as in our main evaluation. 
The neural baseline is trained and evaluated using five repetitions of three-fold cross-validation in order to obtain a stable estimate of the performance and make use of the entire dataset~\citep{BURMAN_1989}.
For each fold, we train the model for fifteen epochs.

The random baseline samples each unidirectional prediction from a distribution derived from the relation distribution of our dataset.
Although our dataset contains no hyponymy or meronymy pairs, these relations are included in the label space for unidirectional prediction.
We therefore divide the probability of hypernymy equally between hypernymy and hyponymy.
Specifically, if the proportion of hypernymy pairs in the dataset is $p$, hypernymy and hyponymy are each assigned a sampling probability of $\frac{p}{2}$. 
The same procedure is applied to holonymy and meronymy.
The probability of antonymy remains unchanged.

\section{Full Results}
\subsection{Instruction Following Rate and Prediction Distribution}
\label{app:pred_dist}
\begin{table*}[ht]
\centering\small
\begin{tabular}{lrrrrrr|rrrrrr}
\toprule
& \multicolumn{6}{c}{Zero-shot}
& \multicolumn{6}{c}{Five-shot} \\
\cmidrule(lr){2-7}
\cmidrule(lr){8-13}
Model
& IFR & HYP & HPO & ANT & HOL & MER
& IFR & HYP & HPO & ANT & HOL & MER \\
\midrule

Tiny Aya (3B)
& 93\% & 29\% & 16\% & 01\% & 00\% & 54\%
& 93\% & 56\% & 14\% & 10\% & 08\% & 12\% \\

Aya Expanse (8B)
& 100\% & 09\% & 27\% & 09\% & 40\% & 16\%
& 100\% & 08\% & 10\% & 14\% & 38\% & 30\% \\

Aya Expanse (32B)
& 100\% & 03\% & 39\% & 28\% & 08\% & 22\%
& 100\% & 04\% & 27\% & 14\% & 15\% & 41\% \\

\midrule

EuroMoE (3B)
& 100\% & 04\% & 02\% & 93\% & 02\% & 00\%
& 100\% & 01\% & 00\% & 99\% & 00\% & 00\% \\

EuroLLM (9B)
& 100\% & 11\% & 11\% & 00\% & 17\% & 60\%
& 100\% & 30\% & 09\% & 09\% & 25\% & 26\% \\

EuroLLM (22B)
& 100\% & 03\% & 83\% & 13\% & 00\% & 01\%
& 100\% & 25\% & 10\% & 35\% & 06\% & 23\% \\

\midrule

Marco-Nano (8B)
& 00\% & 51\% & 28\% & 06\% & 12\% & 03\%
& 100\% & 00\% & 92\% & 05\% & 00\% & 03\% \\

Marco-Mini (17B)
& 68\% & 01\% & 90\% & 09\% & 00\% & 00\%
& 100\% & 32\% & 36\% & 22\% & 01\% & 09\% \\

\midrule

Gemma 4 (12B)
& 19\% & 50\% & 35\% & 07\% & 04\% & 04\%
& 22\% & 04\% & 37\% & 49\% & 03\% & 07\% \\

Gemma 4 (26B)
& 83\% & 37\% & 32\% & 30\% & 00\% & 01\%
& 98\% & 20\% & 40\% & 33\% & 00\% & 07\% \\

Gemma 4 (31B)
& 99\% & 31\% & 51\% & 12\% & 01\% & 05\%
& 100\% & 35\% & 46\% & 09\% & 00\% & 09\% \\

\midrule

Qwen3.5 (9B)
& 100\% & 00\% & 68\% & 29\% & 01\% & 02\%
& 100\% & 14\% & 45\% & 31\% & 02\% & 08\% \\

Qwen3.5 (27B)
& 100\% & 35\% & 39\% & 24\% & 01\% & 02\%
& 100\% & 27\% & 49\% & 13\% & 02\% & 09\% \\

Qwen3.5 (35B)
& 100\% & 43\% & 45\% & 08\% & 01\% & 02\%
& 100\% & 15\% & 56\% & 19\% & 02\% & 09\% \\

Qwen3.5 (122B)
& 100\% & 20\% & 68\% & 10\% & 00\% & 02\%
& 100\% & 18\% & 52\% & 19\% & 04\% & 07\% \\

\midrule

GPT-5.6  Luna
& 100\% & 52\% & 40\% & 06\% & 00\% & 02\%
& 100\% & 23\% & 58\% & 16\% & 00\% & 03\% \\

GPT-5.6  Terra
& 100\% & 32\% & 40\% & 21\% & 03\% & 05\%
& 99\% & 31\% & 36\% & 14\% & 03\% & 17\% \\

\bottomrule
\end{tabular}
\caption{Instruction-following rates and prediction distributions by model and setting. 
IFR stands for instruction-following rate.}
\label{tab:pred_dist}
\end{table*}
In Table~\ref{tab:pred_dist}, we report the prediction distribution and instruction-following rate for each model and setting.
The instruction-following rate~(IFR) refers to the proportion of predictions in which the model outputs an integer from 1 to 5, as instructed in the system prompt.
The prediction distribution is calculated over predictions that follow this instruction.
For example, among the 93\% of predictions from Tiny Aya (3B) that follow the instruction in the zero-shot setting, 29\% are hypernymy.
The IFR of Marco-Nano is shown as 00\% due to rounding, although 224 predictions actually follow the instruction.

\subsection{CCV}
\label{app:ccv_all}
\begin{table}[ht]
    \centering\small
        \begin{tabular}{l|cc}
    \toprule
    Model
    & \multicolumn{1}{c}{Zero-shot}
    & \multicolumn{1}{c}{Five-shot} \\
    \midrule
        Tiny Aya (3B)     & \textbf{48.5} & 15.6 \\
        Aya Expanse (8B)  & 14.4 & 10.2 \\
        Aya Expanse (32B)  & \textbf{31.2} & 12.4 \\
        \midrule 
        EuroMoE (3B) & \phantom{0}8.7 & \phantom{0}3.5\\
        EuroLLM (9B)   & \textbf{31.1} & \textbf{25.2}\\
        EuroLLM (22B)  & \textbf{22.9} & \textbf{29.6}\\
        \midrule
        Marco-Nano (8B) & -- & \textbf{33.6} \\
        Marco-Mini (17B) & \textbf{89.6} & 11.9\\
        \midrule
        Gemma 4 (12B)  & \textbf{45.1} & 51.1\\
        Gemma 4 (26B)  & 18.9 & 14.5\\
        Gemma 4 (31B)  & 9.1 & 9.8\\
        \midrule
        Qwen3.5 (9B)  & 17.9 & 13.8 \\
        Qwen3.5 (27B)  & 12.8 & 10.1\\
        Qwen3.5 (35B)  & 10.3 & 12.8\\
        Qwen3.5 (122B)  & 11.5 & 14.0\\
        \midrule
        GPT-5.6 Luna  & \phantom{0}5.9 & 15.8\\
        GPT-5.6 Terra  & 12.6 & \phantom{0}9.6\\
    \bottomrule
    \end{tabular}
    \caption{
        CCV~(\%) of LLMs by setting. 
        Boldface indicates values above the reference threshold of 20\%.
    }
    \label{tab:ccv_all}
\end{table}
We present the full results of CCV in Table~\ref{tab:ccv_all}.

\section{Licences and Computational Resources}
We use the July 2026 release of Cygnet~\citep{Maudslay_Bond_2026}.
The glosses originate from the English~\citep{Mccrae_2019}, 
Greek~\citep{Stamou_2004}, 
Romanian~\citep{Tufis_2008}, 
Turkish~\citep{Bilgin_2004}, and 
Japanese~\citep{Isahara_2008, Bond_2009} WordNets, released under CC BY 4.0, Apache 2.0, CC BY-SA 4.0, CC BY 4.0, and the Japanese WordNet licence\footnote{\url{https://bond-lab.github.io/wnja/}}, respectively.
Open-source models are used under the licences given on their model cards, and the proprietary models were accessed via the OpenAI API (\texttt{gpt-5.6-luna} and \texttt{gpt-5.6-terra}).
Stanza is released under the Apache 2.0 licence and mE5-large under the MIT licence.
We use all data and models for research purposes, consistent with their intended use.
Our dataset will be released for research on the evaluation of semantic relation knowledge, under a licence compatible with those of the source WordNets.

Inference with open-source models was run with Hugging Face Transformers (version 5.16.0) on a single NVIDIA H100 GPU per model, except for Qwen3.5 (122B), which required four.
All experiments took approximately one week of wall-clock time.
The proprietary models were queried via the OpenAI Batch API in 60 batch jobs~(20 for the zero-shot and 40 for the five-shot setting), comprising approximately 0.6 million requests.
Statistical tests and the Benjamini--Yekutieli correction were conducted with SciPy (version 1.18.1).

\end{document}